\documentclass[a4paper, 10pt, conference]{ieeeconf}      
\usepackage{FG2019}

\FGfinalcopy 

\IEEEoverridecommandlockouts                              
\usepackage{times}
\usepackage{xcolor}
\usepackage{soul}
\usepackage[utf8]{inputenc}
\usepackage[small]{caption}
\usepackage{mathrsfs}
\usepackage{amssymb}
\usepackage{graphicx}
\usepackage{pdfpages}
\usepackage{balance}
\usepackage{booktabs}
\usepackage[flushleft]{threeparttable}
\usepackage{amsmath}

\def\FGPaperID{13} 

\title{\LARGE \bf
Segmentation Guided Image-to-Image Translation\\
with Adversarial Networks
}
\author{\parbox{16cm}{\centering
    {\large Songyao Jiang, Zhiqiang Tao and Yun Fu}\\
    {\normalsize
    Department of Electrical and Computer Engineering, Northeastern University, United States of America\\}}
    \thanks{This work was not supported by any organization.}
}

\begin{document}

\ifFGfinal
\thispagestyle{empty}
\pagestyle{empty}
\else

\pagestyle{plain}
\fi
\maketitle

\begin{abstract}
Recently image-to-image translation has received increasing attention, which aims to map images in one domain to another specific one. Existing methods mainly solve this task via a deep generative model, and focus on exploring the relationship between different domains. However, these methods neglect to utilize higher-level and instance-specific information to guide the training process, leading to a great deal of unrealistic generated images of low quality. Existing methods also lack of spatial controllability during translation. To address these challenge, we propose a novel Segmentation Guided Generative Adversarial Networks (SGGAN), which leverages semantic segmentation to further boost the generation performance and provide spatial mapping. In particular, a segmentor network is designed to impose semantic information on the generated images. Experimental results on multi-domain face image translation task empirically demonstrate our ability of the spatial modification and our superiority in image quality over several state-of-the-art methods.
\end{abstract}

\section{Introduction}
Image-to-image translation aims to map an image in a source domain to its corresponding image in a target domain~\cite{liu2017unsupervised}, which in essence generalizes a wide range of computer vision and graphics tasks, such as image super-resolutions~\cite{ledig2016photo} (low-resolution to high-resolution), semantic segmentation~\cite{luc2016semantic} (image to semantics), style transfer~\cite{johnson2016perceptual} (photo to paint), and face recognition~\cite{yang2018identity}. Among these interesting topics, face image translation~\cite{kaneko2017generative} draws increasing attentions, where \emph{domain} denotes face images with the same attribute (\emph{e.g.}, hair color, gender, age, and facial expressions) and the task is to change the attributes for a given face image.



Recently, generative adversarial networks (GAN)~\cite{goodfellow2014generative} emerges as a powerful tool for generative tasks, and significantly thrives the field of deep generative models. As GAN could provide realistic image generation results and alleviate the deficiency of training data, a great deal of research efforts~\cite{liu2017unsupervised,zhu2017unpaired,kim2017learning,yi2017dualgan} have been made to tackle image translation with GAN based frameworks. These methods generally devise a \emph{generator} to generate images belonging to a target domain upon the input of images in a source domain , and develop a \emph{discriminator} to distinguish between the generated images (\emph{fake samples}) and the real ones (\emph{real samples}). By leveraging an adversarial training scheme~\cite{goodfellow2014generative}, the discriminator effectively supervises the training of generator, and eventually delivers reliable results.

However, though these GAN-based methods have achieved appealing progress, there still remains two challenges for the image translation task. First, the reliability of GAN based methods are still quite low, which inevitably limits the capability and flexibility for their applications. It is because that previous methods mainly focus on exploring the underlying relationship between different domains, yet neglect to utilize the rich information inside images to further boost the translating performance. Specifically, they only employ the discriminator to supervise generator to capture the distribution of target domain, but ignore the information on instance-level (\emph{e.g.}, facial semantic segmentation) to ensure the image quality. This may badly lower the generation ability, and lead to unrealistic images, such as the notorious ``ghost'' faces. Second, since their training process is built on domain-level labels without strong spatial regulation, existing methods lack the controllability of achieving gradually morphing effects such as changing face shapes, orientations and facial expressions.

\begin{figure}
  \centering
  \includegraphics[width=0.46\textwidth]{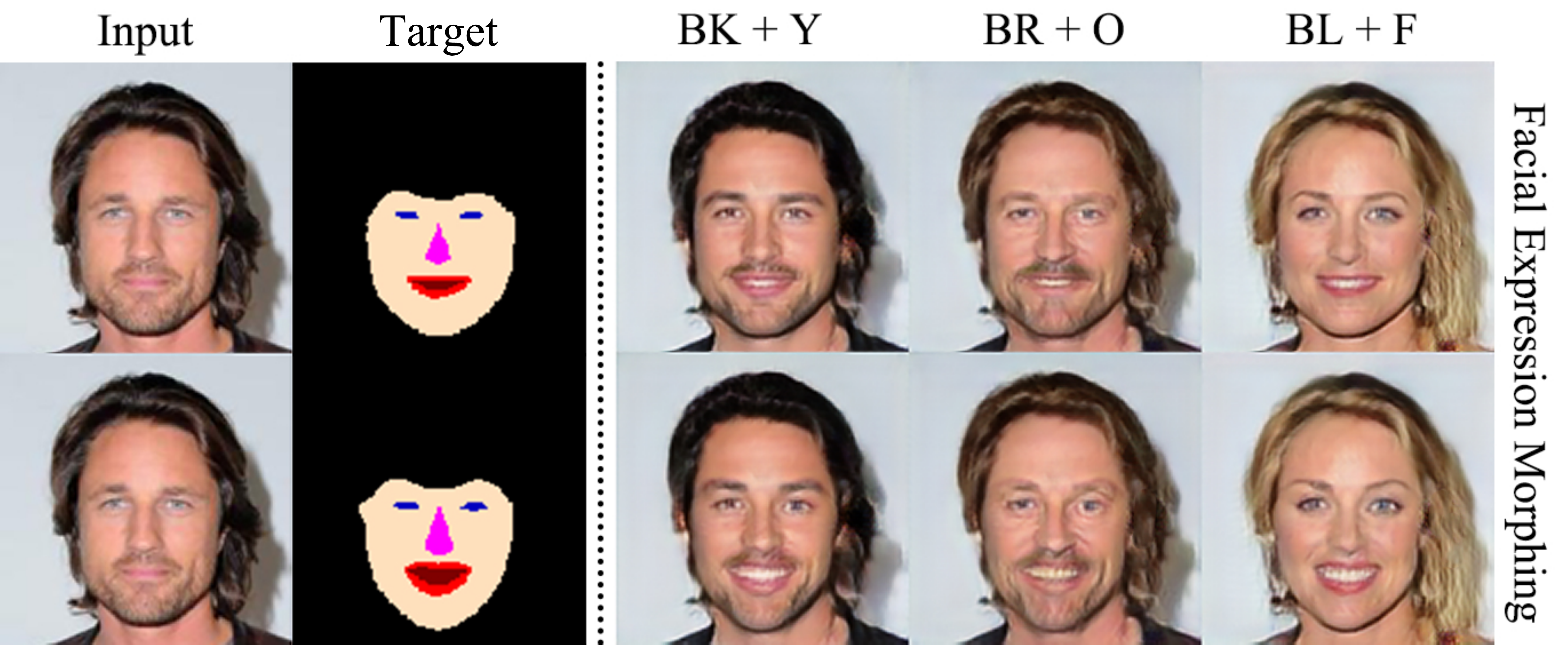}
  \caption{Given an input image with a target segmentation, the proposed SGGAN can translate image to various combination of target facial attributes as well as morphing face into the target expression. Abbreviation: BK=Black Hair, Y=Young, O=Old, F=Female.}\label{fig_cover}\vspace{-0.6cm}
\end{figure}

\begin{figure*}[t]
  \centering
  \includegraphics[width=0.9\textwidth]{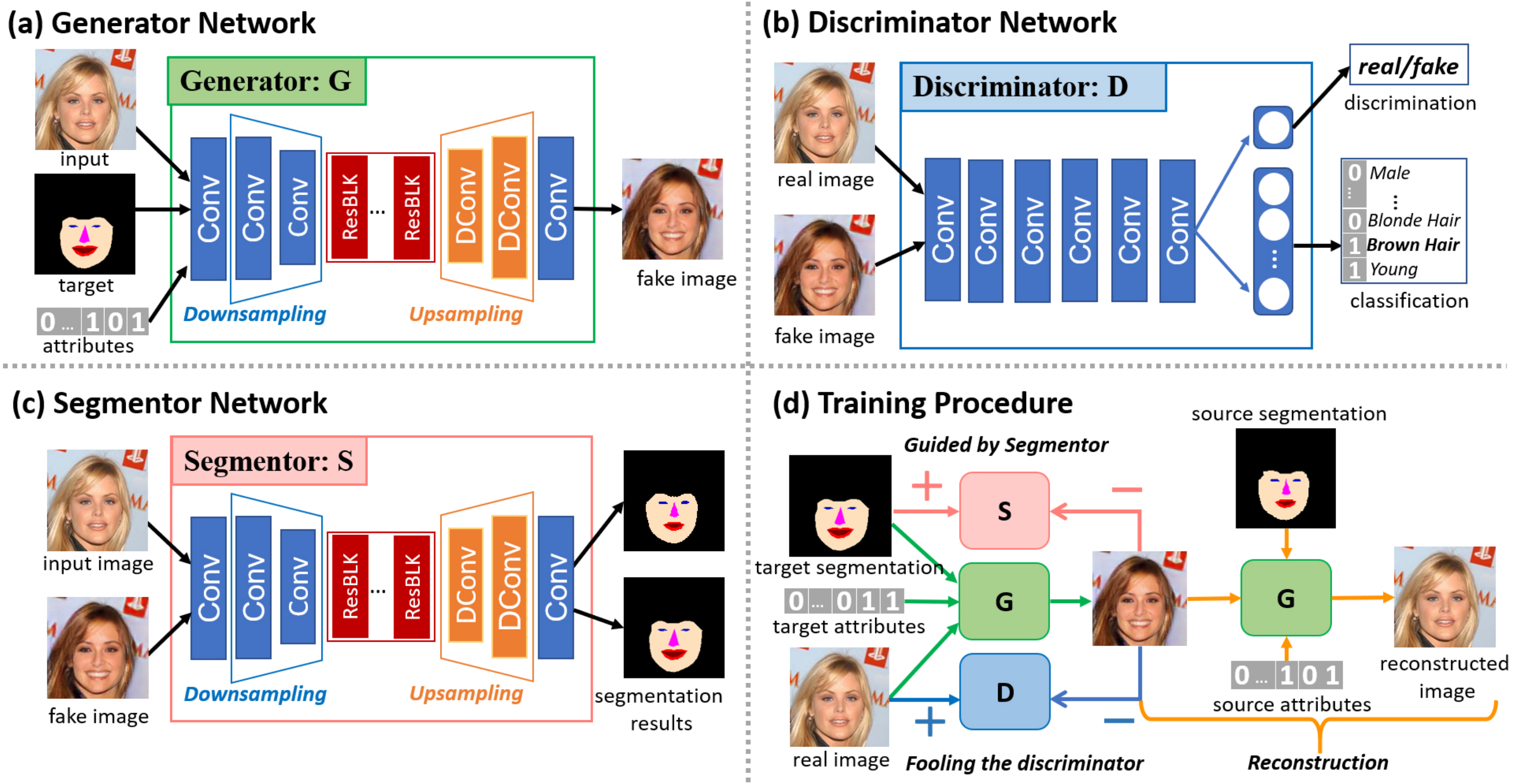}
  \caption{Illustration of SGGAN. SGGAN contains (a) a generator $G$, (b) a discriminator with auxiliary classifier $D$ and (c) a segmentor $S$. During training, $S$ provides spatial guidance to $G$ to ensure the generated images comply with input segmentations. $D$ aims to ensure the translated images are realistic as the real images. The training procedure of the whole framework is illustrated in (d). }\label{fig_model}\vspace{-0.3cm}
\end{figure*}

To address the above challenges, we propose a novel Segmentation Guided Generative Adversarial Networks (SGGAN), which fully leverages segmentation information\footnote{
  The image semantic segmentation could be obtained through multiple ways, such as human annotations or given by any segmentation algorithms. Here, however, we focus on face image generation and thus provide an unsupervised way to obtain semantic segmentation based on extracted facial landmarks.
} to guide the image translation process. In detail, as illustrated in Fig. \ref{fig_model}, the proposed SGGAN framework consists of three networks, \textit{i.e.}, a generator, a discriminator and a segmentor: (1) The generator takes as inputs a given image, multiple attributes and a target segmentation to generate a target image which is expected to be consistent with the input attributes and segmentation; (2) The discriminator pushes the generated images towards the target domain distribution, and meanwhile utilizes an auxiliary attribute classifier to enable SGGAN to generate images with target attributes; (3) The segmentor focuses on a segmentation task where it is fed with either a real image or a fake image, and generate its corresponding estimated semantic segmentations. During training SGGAN, the estimated segmentations from the segmentor are compared with their ground-truth values, which provides gradient information to optimize the generator network. This optimization tends to teach the generator to imposes the spatial constraints extracted from the semantic segmentation on the translated images. The benefits of introducing the segmentor network lies at two folds. First, it explicitly guides the generator with pixel-level semantic segmentations, and thus further boosts the image quality. Second, the target segmentation works as a strong prior for the generator, which could use this prior to edit the spatial content and align the face image to the target segmentation. By this means, our approach can simultaneously change facial attributes and achieve facial expression morphing without giving extra expression labels, as shown in Fig~\ref{fig_cover}.

In this paper, extensive experiments on several face image translation tasks are presented to empirically demonstrate the effectiveness of our proposed SGGAN, compared with several state-of-the-art image translation methods. We also show that our approach can spatially control the translation process and provide interpretable results. We summarize our contributions as follows.
\begin{itemize}
\item A novel Segmentation Guided Generative Adversarial Networks (SGGAN) model is proposed, which leverages semantic segmentation information to provide spatial constraints for the image translation task.

\item A segmentor network is particularly designed to impose the target spatial guidance on the generator.

\item We provide a general model for face synthesis task, which could generate face images with multi-domain attributes and also handle facial expression morphing.


\end{itemize}

\section{Related work}
In recent years, GAN based methods have become popular and achieved great success in many computer vision tasks such as image super-resolution~\cite{ledig2016photo}, semantic segmentation~\cite{luc2016semantic}, object detection~\cite{li2017perceptual}, video prediction~\cite{mathieu2015deep} and classification~\cite{yu2017open,li2017end}. Many research efforts are conducted to improve GAN in stablizing the training process and increasing the generated diversity~\cite{zhao2016energy,mao2017least,arjovsky2017wasserstein,berthelot2017began,gulrajani2017improved}, improving the visual quality and resolution of generated images~\cite{wang2017high,karras2018progressive}, introducing controllability by adding conditional label information~\cite{mirza2014conditional,li2017triple}, and increasing its interpretability~\cite{chen2016infogan,kaneko2017generative}.

Recently, \cite{isola2016image} propose an image-to-image translation networks called pix2pix which uses an image as the conditional input and train their networks supervisedly with paired image data. Many researchers then find that supervision is not necessary for image-to-image translation task and develop their unsupervised methods such as CycleGAN~\cite{zhu2017unpaired}, DiscoGAN~\cite{kim2017learning}, DualGAN~\cite{yi2017dualgan} and UNIT\cite{liu2017unsupervised}. These methods are essentially two-domain image translation methods which translate images from their source domain to a target domain using one-to-one mappings.

Based on their work, \cite{choi2017stargan} propose a multi-domain image-to-image translation framework called StarGAN, which utilizes an auxiliary classifier following \cite{odena2016conditional} to achieve a one-to-many mapping between a source domain and multiple target domains. But their method still may generate unrealistic low-quality output images, and lacks of spatial controllability. Different from \cite{choi2017stargan}, our proposed SGGAN framework introduces instance-level target segmentation as strong regulations to guide the translating process, which avoids fake flaws and makes the translated results spatially controllable.

\section{Methodology}
In this section, we first give the problem formulation to our method, then elaborate the proposed segmentor network, and finally give the overall objective function of our model.

\subsection{Problem Formulation}
Let $x$, $s$ and $c$ be an image of size ($H \times W \times 3$), a segmentation map ($H \times W \times n_{s}$) and an attribute vector ($1 \times n_{c}$) in the source domain; while $y$, $s'$ and $c'$ be their corresponding image, segmentation and attributes in the target domain. We denote $n_{s}$ as the number of segmentation class, and $n_{c}$ as the number of all the attributes. Note that, for $s$ and $s'$, each pixel is represented by a one-hot vector of $n_{s}$ classes, while for $c$ and $c'$, they are binary vectors of multiple labels, since we consider the scenario of multi-domain translation. Thus, in this paper, our goal is to find such a mapping that $G\left (x,s',c'\right ) \rightarrow y$.

To achieve this, as illustrated in Fig. \ref{fig_model}, we formulate $G$ as the generator network in our proposed SGGAN model. Meanwhile, we employ a discriminator network $D$ and a segmentor network $S$ to supervise the training of $G$. As following~\cite{choi2017stargan}, $D$ is developed with two different purposes to handle multi-attribute labels, such as $D: x \rightarrow\{D_{a}, D_{c}\}$. In details, $D_{a}(\cdot)$ outputs a single scalar that represents the probability of the given sample belonging to the target domain, and $D_{c}(\cdot)$ gives a vector of size ($1 \times n_c$) with each dimension being the probability of one specific attribute.

\subsection{Segmentor Network}
In order to guide the generator by the target segmentation, we build an additional network which takes an image as input and generate its corresponding semantic segmentation. We refer to this network as the segmentor $S$ which is trained together with the GAN framework. As illustrated in Fig.~\ref{fig_model}(a) and (d), when training with the real data pairs ($x,s$), $S$ learns to estimate segmentation correctly. When $S$ is trained together with $G$, the fake image denoted by $G\left(x,s',c'\right)$ is fed to $S$ to obtain its estimated segmentation $S\left(G\left(x,s',c'\right)\right)$, which is compared with $s'$ to calculate a segmentation loss. When optimizing $G$, with minimizing the segmentation loss providing gradient information, $G$ tends to translate the input image to be consistent with $s'$. To better utilize the information in $s'$, $s'$ is annotated as a $k$-channel image that each pixel is represented by a one-hot vector indicating its class index. Then $s'$ is concatenated to $x$ in channel dimension before feeding into the generator. In summary, we leverages semantic segmentation information in GAN based image translation tasks and we also build a segmentor which is trained together with GAN framework to provide guidance in image translation.

\begin{figure}[t]
    \centering
    \includegraphics[width=0.48\textwidth]{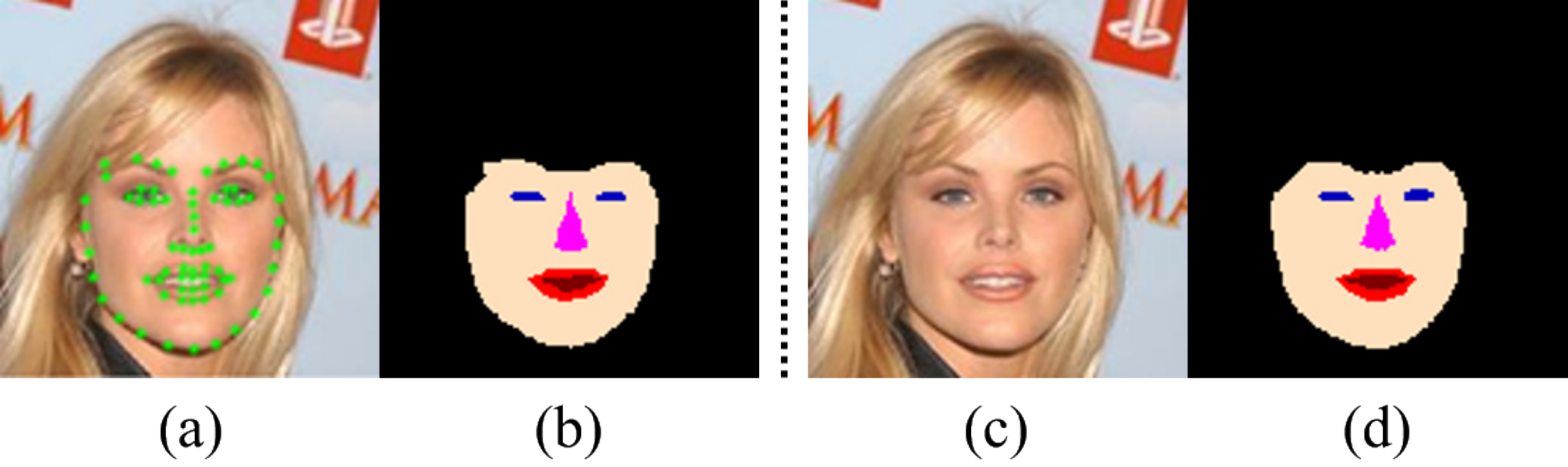}
    \caption{Illustration of segmentor network. (a) Facial landmarks extracted. (b) Landmarks based semantic segmentation. (c) Real image sample. (d) Segmentor generated segmentation from (c).}
    \label{fig_segmentor}
\end{figure}

Here we introduce our techniques to obtain semantic segmentations of face images and train the segmentor. As illustrated in Fig. \ref{fig_segmentor}, a great number of face alignment methods can be applied to extract the facial landmarks \ref{fig_segmentor}(a) from an input image \ref{fig_segmentor}(c). We then process extracted landmarks to generate a pixel-wised semantic segmentation as shown in \ref{fig_segmentor}(b) that each pixel in the input image is automatically classified into classes of eyes, eyebrow, nose, lips, skin and background according to landmarks information. In training phase, we takes a real image sample \ref{fig_segmentor}(c) as an input to $S$ and generate its estimated segmentation \ref{fig_segmentor}(d). We optimize $S$ by minimizing the difference between \ref{fig_segmentor}(b) and \ref{fig_segmentor}(d).

\subsection{Optimization of SCGAN}
With the segmentor presented above, we propose SGGAN, which utilize semantic segmentations as strong regulations and control signals in multi-domain image-to-image translation. In this subsection, we introduce the loss functions to optimize those networks and define their purposes.

\noindent\textbf{Segmentation Loss. }
To regulate the generated face image to comply with the target segmentation, we propose a segmentation loss, which acts as an additional regulation and guides the generator to generate target fake images. Taking a real image sample $x$ as input, the generated segmentation $S\left(x\right)$ is compared with the source segmentation $s$ to optimize the segmentor $S$. The loss function can be described as
\begin{equation}
\label{seg_loss_real}
\mathcal{L}_{seg}^{real} = \mathbb{E}_{x,s}[A_s(s, S(x)],
\end{equation}
where $A_s(\cdot, \cdot)$ computes cross-entropy loss pixel-wisely by
\begin{equation}
A_s(a, b) = -\sum_{i=1}^{H}\sum_{j=1}^{W}\sum_{k=1}^{n_s} a_{i,j,k} \log b_{i,j,k},
\end{equation}
with $a, b$ being two segmentation maps of size ($H \times W \times \ n_s$).

To guide the generator to generate desired target images, a generated image is fed into $S$ to get a fake segmentation which is compared with the target segmentation input to the $G$. This fake segmentation loss could guide $G$ to generate images comply with target segmentation, which is defined as
\begin{equation}
\label{seg_loss_fake}
\mathcal{L}_{seg}^{fake}=\mathbb{E}_{x,s',c'}\left[A_s(s', S(G(x, s', c')))\right].
\end{equation}

\begin{figure*}[t]
  \centering
  \includegraphics[width=0.97\textwidth]{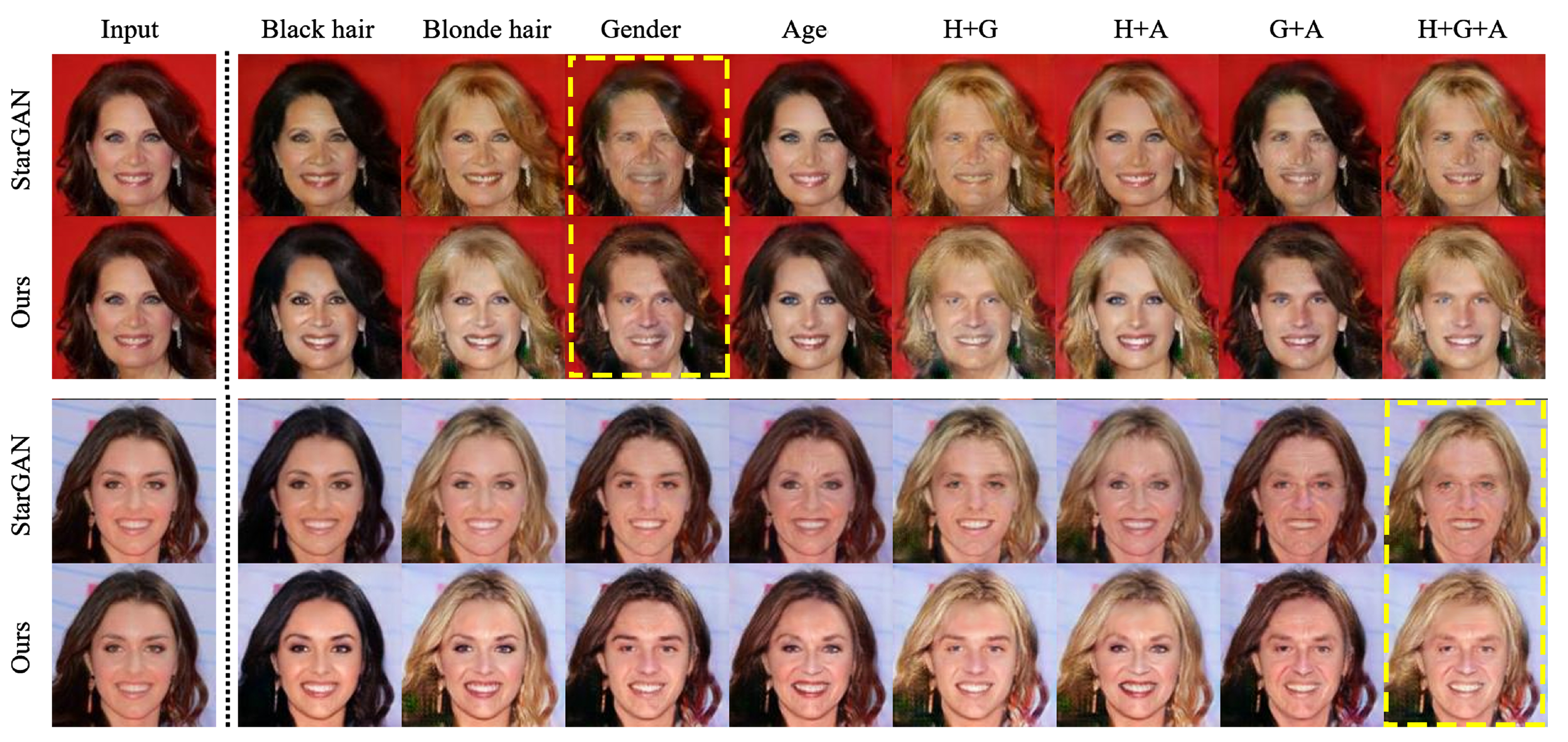}
  \caption{Multi-domain face translation results compared with StarGAN. The input images are shown in the first column. Same combinations of target attributes to be translated are selected as inputs. Yellow rectangle highlight our major improvement over StarGAN. We use abbreviations to denote the names of attributes: H=Hair color, G=Gender, A=Age.}\label{fig_compare_stargan}\vspace{-0.3cm}
\end{figure*}

\noindent\textbf{Adversarial Loss. }
The proposed SGGAN generates two types of images. The first one is the fake image generated by $G$ from input image with target segmentation and target attributes denoted by $G\left(x,s',c'\right)$. The second one is the reconstructed image generated from fake images, source segmentation and source labels represented by $G\left(G\left(x,s',c'\right),s,c\right)$. We adopt an adversarial loss to the former path and thus form a generative adversarial networks with the discriminator $D$. The later path reconstructs the input image in the source domain using the fake image, which can be trained with supervision using input image that additional adversarial loss is unnecessary. The adversarial loss is defined as

\begin{equation}\label{adv_loss}
\begin{aligned}
\mathcal{L}_{adv}=
 & \mathbb{E}_{x}\left[\log D_{a}\left(x\right)\right] + \\
 &\mathbb{E}_{x,s',c'}\left[\log \left(1-D_{a}\left(G\left( x,s',c'\right)\right)\right)\right].
\end{aligned}
\end{equation}

By optimizing the adversarial loss, $G$ tends to generate face images which can not be distinguished from real images.

\noindent\textbf{Classification Loss. }
In order to obtain attribute-level domain translation ability, we implement an auxiliary attributes classifier $A_c$, which shares weights with $D$ except output layer as following \cite{choi2017stargan}. $A_c$ acts like a multi-class classifier which classifies the face image to their attributes labels.
Objective functions associated with $A_c$ contains one loss for real image $x$ to train the classifier which is defined as
\begin{equation}
\label{cls_loss_real}
\mathcal{L}_{cls}^{real}=\mathbb{E}_{x, c} \left[ A_c(c, D_{c}(x) ) \right],
\end{equation}
where $A_c(\cdot, \cdot)$ computes a multi-class cross-entropy loss by $A_c(a, b) = -\sum_{k}a_k\log(b_k)$ with $a,b$ being two vectors of identical size ($1 \times n_c$). Accordingly, we have the $\mathcal{L}_{cls}^{fake}$ for generated fake images by
\begin{equation}
\label{cls_loss_fake}
\mathcal{L}_{cls}^{fake}=\mathbb{E}_{x, s', c'} \left[ A_c(c', D_{c}( G(x, s', c') ) )\right].
\end{equation}


\noindent\textbf{Reconstruction Loss. }
We also adopt a reconstructive cycle which translates $x$ into its corresponding target domain $\left(s',c'\right)$, then translates back into the source domain $(s,c)$. This loss aims to keep the basic contents of $x$ during image translation. In this path, the reconstructed image $G\left(G\left(s',c'\right),s,c\right)$ should be as close as $x$. The reconstruction loss is defined as
\begin{equation}
\label{rec_loss}
\mathcal{L}_{rec}=\mathbb{E}_{x, s', c', s, c}\left[\left\|x-G\left(G(x, s',c'),s,c\right)\right\|_1\right].
\end{equation}

\noindent\textbf{Overall Objective. }
Full objective function of our SGGAN network to optimize $G$, $D$ and $S$ could be summarized as
\begin{equation}
\label{full_loss_S}
\mathcal{L}_{S}=\mathcal{L}_{seg}^{real},
\end{equation}
\begin{equation}
\label{full_loss_D}
\mathcal{L}_{D}=-\mathcal{L}_{adv}+\lambda_{1}\mathcal{L}_{cls}^{real},
\end{equation}
\begin{equation}
\label{full_loss_G}
\mathcal{L}_{G}=\mathcal{L}_{adv}+\lambda_{1}\mathcal{L}_{cls}^{fake}+\lambda_{2}\mathcal{L}_{seg}^{fake}+\lambda_{3}\mathcal{L}_{rec}
\end{equation}
where $\lambda_{1}$, $\lambda_{2}$ and $\lambda_{3}$ are hyper-parameters which control the weights of classification loss, segmentation loss and reconstruction loss. These weights act as relatively importance of those terms compared to adversarial loss. Since $A_c$ is embedded in $D$ and shares the same weights except the output layer, $A_c$ is trained together with $D$ using discriminator loss $\mathcal{L}_{D}$ which contains both the adversarial term and the classification term on real image samples.

\noindent\textbf{Training and Testing. }
In the training phase, a batch of ($x,s,c$) are samples from the real data distribution. Their target $s'$ and $c'$ are obtained by randomly shuffling $s$ and $c$. The SGGAN can then be optimized using their objective functions. In the testing phase, when doing the attribute translation only, we use the trained $S$ to obtain the segmentation of the test image $x$ as the target $s'$ to keep the spatial contents unchanged. When we do the spatial translation, we select any desired $s'$ from the dataset. $G$ can then align $x$ to the target $s'$. In our experiment, we use $\lambda_{1}=1$, $\lambda_{2}=10$ and $\lambda_{3}=5$.


\begin{figure*}[ht]
  \centering
  \includegraphics[width=0.98\textwidth]{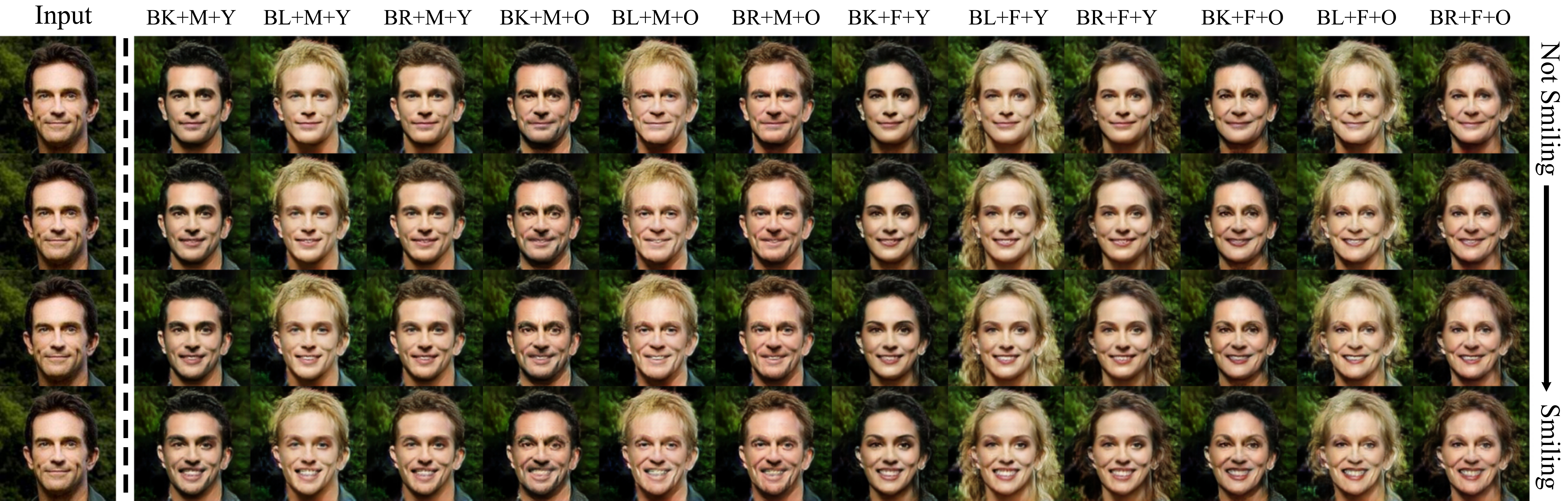}
  \caption{Multiple facial attributes translation together with Nosmile2Smile Interpolation on CelebA dataset (better viewed on screen). The first column shows the input images. The other columns show the results of multi-attributes translation. In total, there are 12 kinds of target attribute translations which are all combinations of selected attributes: hair colors, genders and ages. Meanwhile, each row represents interpolation results between not smiling and smiling faces based on multi-attributes translation results. Abbreviations are used to denote multi-attribute labels: BK=Black hair, BL=Blonde hair, BR=Brown hair, M=Male, F=Female, Y=Young, O=Old.}\label{fig_all}
\end{figure*}

\begin{table}
\caption{Network architecture for SGGAN. CONV=Convolution layer, DCONV=Transposed convolution layer, RESBLK=Residual block, N=Number of filters, K=Kernel size, S=Stride, P=Padding. IN=Instance Normalization, lRELU=leaky RELU activation.}\label{tab_network_architecture}
\centering
\scalebox{0.85}{
\centering
\begin{tabular}{ll}
  \toprule
  Architecture-A & Architecture-B\\
  \toprule
  CONV-(N64 ,K7,S1,P3),IN,RELU & CONV-(N64 ,K4,S2,P1),IN,lRELU \\
  CONV-(N128,K4,S2,P1),IN,RELU & CONV-(N128,K4,S2,P1),IN,lRELU \\
  CONV-(N256,K4,S2,P1),IN,RELU & CONV-(N256,K4,S2,P1),IN,lRELU \\
  RESBLK-(N256,K3,S1,P1),IN,RELU $\times k$ & CONV-(N512,K4,S2,P1),IN,lRELU\\
  DCONV-(N128,K4,S2,P1),IN,RELU & CONV-(N1024,K4,S2,P1),IN,lRELU \\
  DCONV-(N64 ,K4,S2,P1),IN,RELU & CONV-(N2048,K4,S2,P1),IN,lRELU \\
  \midrule
  \multicolumn{2}{l}{Generator: Architecture-A + CONV-(N3,K7,S1,P3),TanH}\\
  \multicolumn{2}{l}{Discriminator: Architecture-B + CONV-(N1,K3,S1,P1) \& CONV-($n_c$,K2,S1,P1)}\\
  \multicolumn{2}{l}{Segmentor: Architecture-A + CONV-($n_s$,K7,S1,P3)}\\
  \bottomrule
\end{tabular}}
\end{table}

\section{Experiment}
In the experiment, SGGAN is compared with recent methods on two-domain and multi-domain face image translations. Then we show our capability of transferring facial attribute and morphing facial expression with a single model.

\subsection{Settings}
\noindent\textbf{Dataset. }
CelebA dataset \cite{liu2015faceattributes} contains 202,599 face images of celebrities with 40 binary attributes labels such as gender, age and hair color which is ideal for multi-domain translation task. We separate this dataset into training and testing data. We use aligned images, crop the center region and resize them to $128\times 128$ in all of our experiments. Facial landmarks detector from Dlib \cite{dlib09} is used to extract landmarks. Since the detector may fail and return invalid results, we remove the failed detection by comparing the detected 68-point landmarks with the ground-truth 5-point landmarks in data preprocessing. Based on extracted 68-point landmarks, we generate semantic facial segmentations consist of eyes, nose, mouth, skin and background regions.


\noindent\textbf{Compared methods. }In our experiment, we compare our results with two-domain translation model CycleGAN\cite{zhu2017unpaired}, UNIT\cite{liu2017unsupervised} and multi-domain translation model StarGAN\cite{choi2017stargan} which represent the state-of-art work in image-to-image translation. Since there are no available pre-trained model, we retrain there models using their published source code. In order to obtain a fair comparison, We train their model on the same dataset with the same number of epochs using their default configurations.

\noindent\textbf{Implementation details. }
The network architecture of SGGAN is shown in Table \ref{tab_network_architecture}. We employ a deep encoder-decoder architecture for both $G$ and $D$ with several residual blocks to increase the depth of our network while avoiding gradients vanishing. For the discriminator, we adopt state-of-the-art loss function and training procedures from improved WGAN with gradient penalty \cite{gulrajani2017improved} to stabilize the training process. In bottleneck layers, $k=6$ residual blocks are implemented for $G$ and $k=4$ residual blocks for the $S$. We use three Adam optimizers with $beta1$ of 0.5 and $beta2$ of 0.999 to optimize our networks. The learning rates are set to be 0.0001 for both $G$ and $D$ and 0.0002 for $S$.

\begin{figure}[t]
  \includegraphics[width=0.48\textwidth]{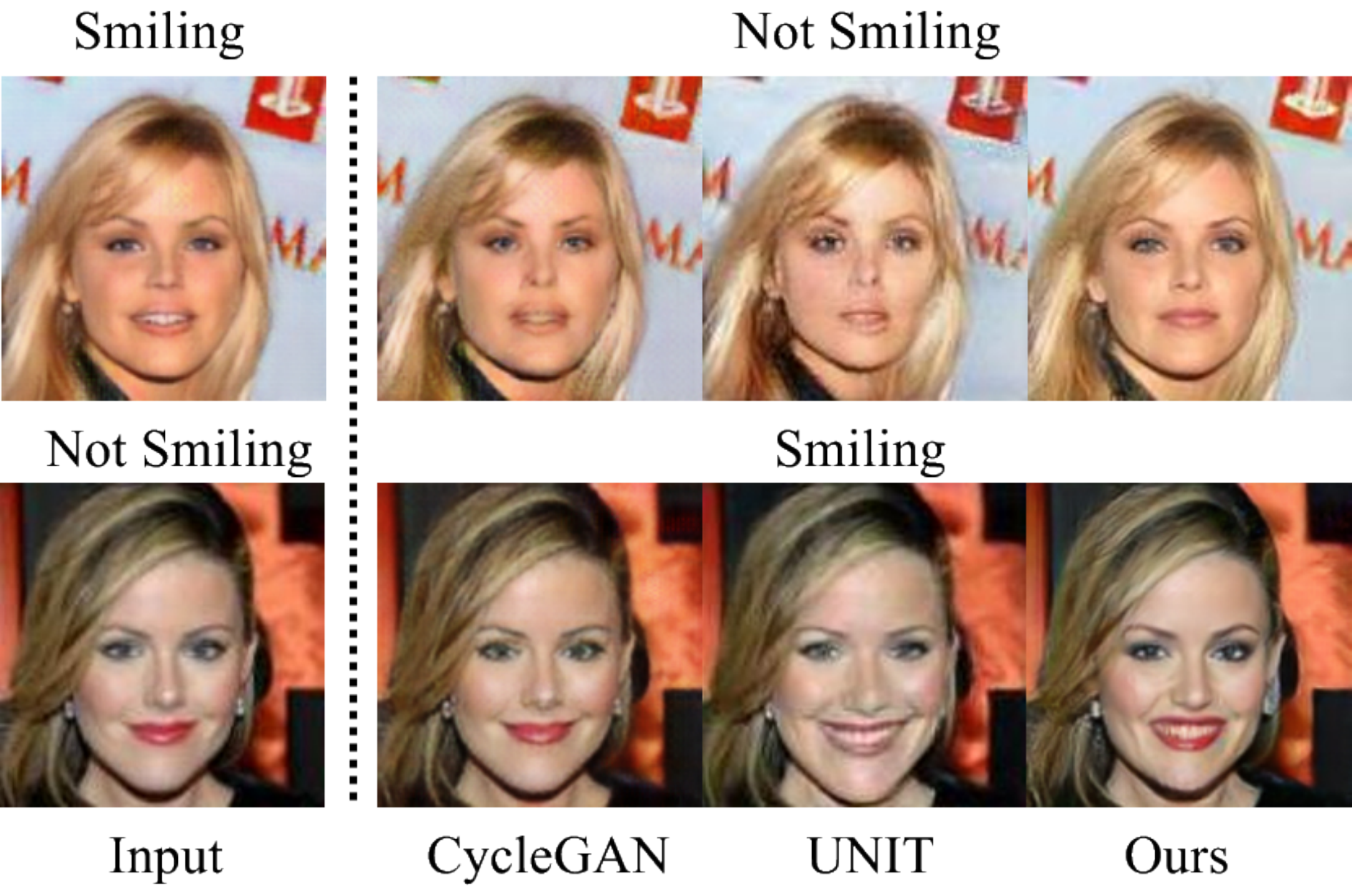}
  \caption{The first row shows the  \textit{Smile2Nosmile} translation results and the second row shows the \textit{NoSmile2Smile} translation results compared with CycleGAN and UNIT.}\label{fig_compare_cyclegan}
  \vspace{-0.2cm}
\end{figure}

\subsection{Image Translations}
\noindent\textbf{Multi-domain image translation. }
In multi-domain image translation task, SGGAN is trained on CelebA dataset with both facial segmentations and attribute-level labels. For fair comparison, we follow the choice of attributes the same as StarGAN \cite{choi2017stargan} in their paper, which are hair color, gender, age and their combinations.

As shown in Fig. \ref{fig_compare_stargan}, SGGAN generally produces much sharper and more realistic results with better contrast compared to StarGAN. It can be seen that, StarGAN does not perform well in transferring gender since their results appear to be vogue especially in multi-attribute transfer tasks. In the meanwhile, their results introduce many unrealistic fake details in the eye and mouth regions, especially when transfer the face from young to old. StarGAN also suffers from a problem that gender-transferring results are too neutral to be regarded as the target gender. In contrary, with guidance of semantic segmentation information, SGGAN effectively transfers all the attributes and produce much sharper, clearer and more realistic translation results, which are considered as our major advantages over StarGAN.

\noindent\textbf{Multi-domain translation with expression morphing. }
In this task, SGGAN model is trained with both segmentation and attributes information. As a result, there are two translating dimensions which are the attributes transfer and the \textit{NoSmile2Smile} interpolation. In attributes transfer, we apply all possible combination of selected attributes which are hair color (black, blonde or brown), gender (male or female) and age (young or old) as our target attributes. In \textit{NoSmile2Smile} interpolation, four-stage morphing segmentations between not smiling face and smiling faces are fed into generator. Fig. \ref{fig_all} shows that our networks can effectively transfer an input image into its realistic target image with specified attributes, warp the face according to the target segmentation, and gradually change facial expressions. From the figure, we can see that introducing a strong regulation provided by facial landmarks based semantic segmentation to guides the generator shows its effectiveness of controlling the spatial contents of the translated face images.

\begin{figure}[t]
\centering
\includegraphics[width=0.48\textwidth]{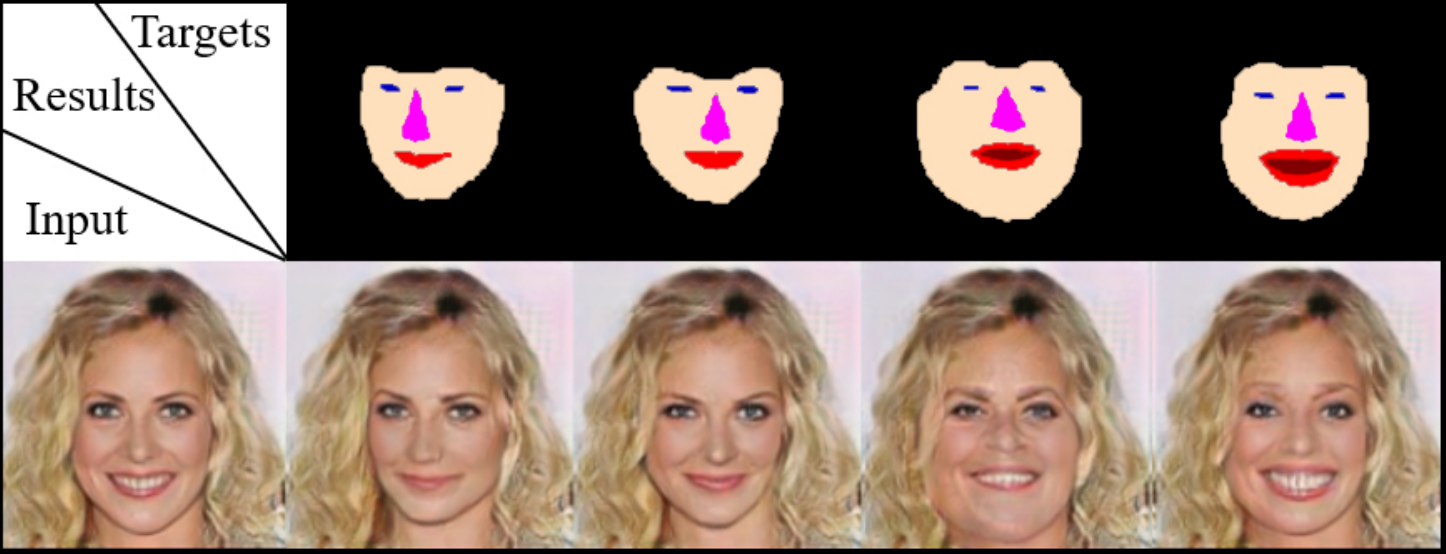}
\caption{Face morphing results. The left-most image is the input image. The first row contains the target segmentations input to the generator. The rest images are the face morphing results.}\label{fig_face_morphing}\vspace{-0.2cm}
\end{figure}

\noindent\textbf{Bi-directional two-domain translation. }
We also compare SGGAN with two-domain translation methods CycleGAN \cite{zhu2017unpaired} and UNIT \cite{radford2015unsupervised} in transferring facial expression bi-directionally between not smiling and smiling. SGGAN is trained on CelebA dataset with segmentation information but without any attribute-level label. CycleGAN and UNIT are trained on CelebA dataset with two images subsets separated by smiling labels. As shown in Fig.~\ref{fig_compare_cyclegan}, in smiling-to-not-smiling direction, both UNIT's and CycleGAN's results look like completely different persons with distorted face shape and blurry details on nose and mouth region which make their facial expressions strange. In not-smiling-to-smiling direction, the result of CycleGAN remains unchanged. We consider this as a result of their identity loss which tends to keep the small smile and disables their ability of further enlarging it. UNIT can successfully enlarge the smile, but with blurry details and fake texture the result are far from good quality. In contrary, in both directions, our results with sharper details and unchanged facial identities are regarded as much more natural and realistic results.

\noindent\textbf{Face morphing (with an ablation study).} We would like to show the power of SGGAN in spatially translating images. Also as an ablation study, we remove the auxiliary attribute classifier from the proposed SGGAN, it is still capable of aligning the input images according to the input target segmentations, which is referred as face morphing in this paper. As shown in Fig.~\ref{fig_face_morphing}, when input a face image with target segmentations of any face shape, facial expression and orientation, our SGGAN can generate faces with target spatial configuration, yet still shares the same attributes with the input face image such as gender, hair color, skin color and background, which demonstrate the effectiveness of the guidance by the target segmentations.

\begin{figure}[t]
\centering
\includegraphics[width=0.48\textwidth]{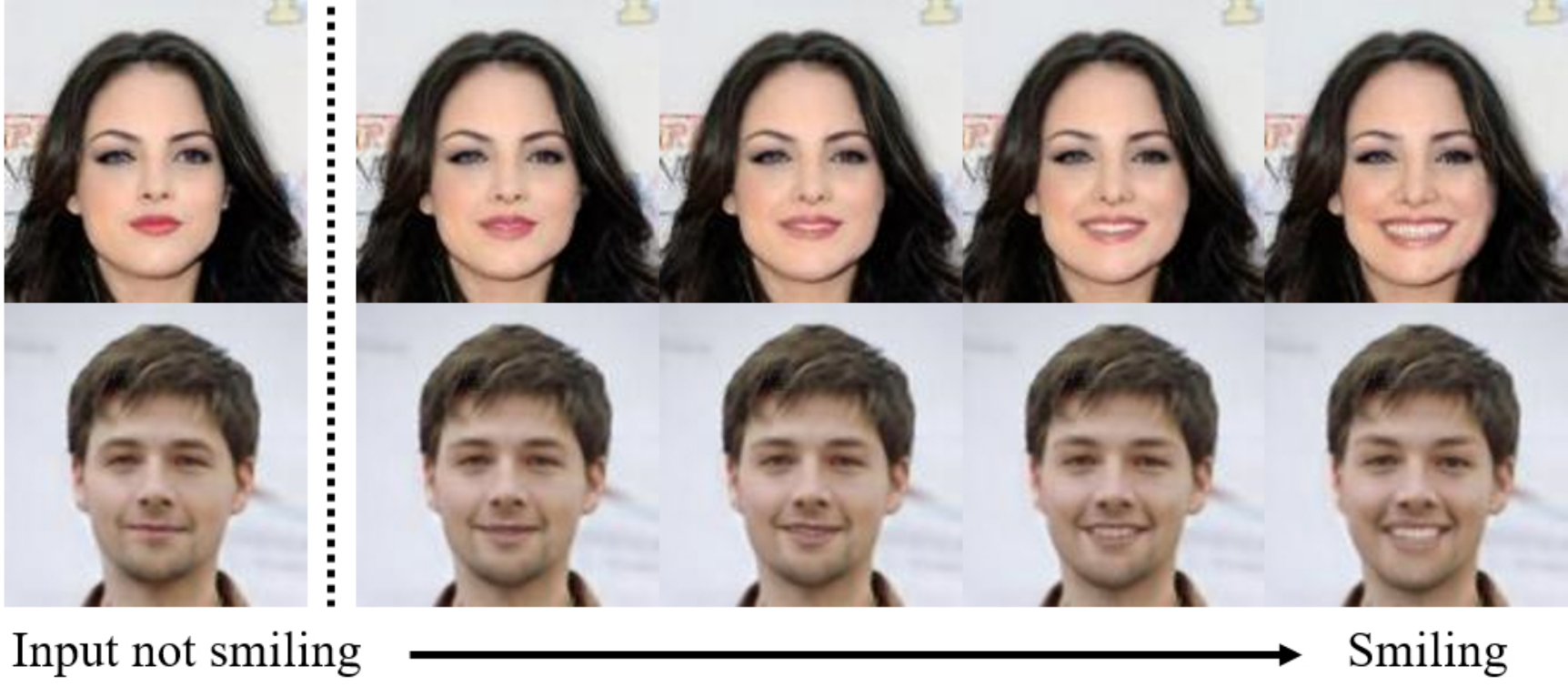}
\caption{\emph{NoSmile2Smile} interpolation. The first column are input images. The right-most column are our translated smiling result. The rest of columns show intermediate results between not smiling and smiling by interpolating on the landmarks.}\label{fig_compare_nosmile2smile_interpolate}
\end{figure}

\subsection{Model Discussion}

\noindent\textbf{Capability of interpolation.~}Moreover, by interpolating facial landmarks points from not-smiling landmarks to smiling landmarks and generating corresponding segmentations, SGGAN could generate intermediate stages between not smiling and smiling expressions as shown in Fig.~\ref{fig_compare_nosmile2smile_interpolate}. All the intermediate results have a good visual quality. Other methods which are trained on binary attribute labels can not achieve the same interpolation results.

\begin{figure}[t]
  \centering
  \includegraphics[width=0.48\textwidth]{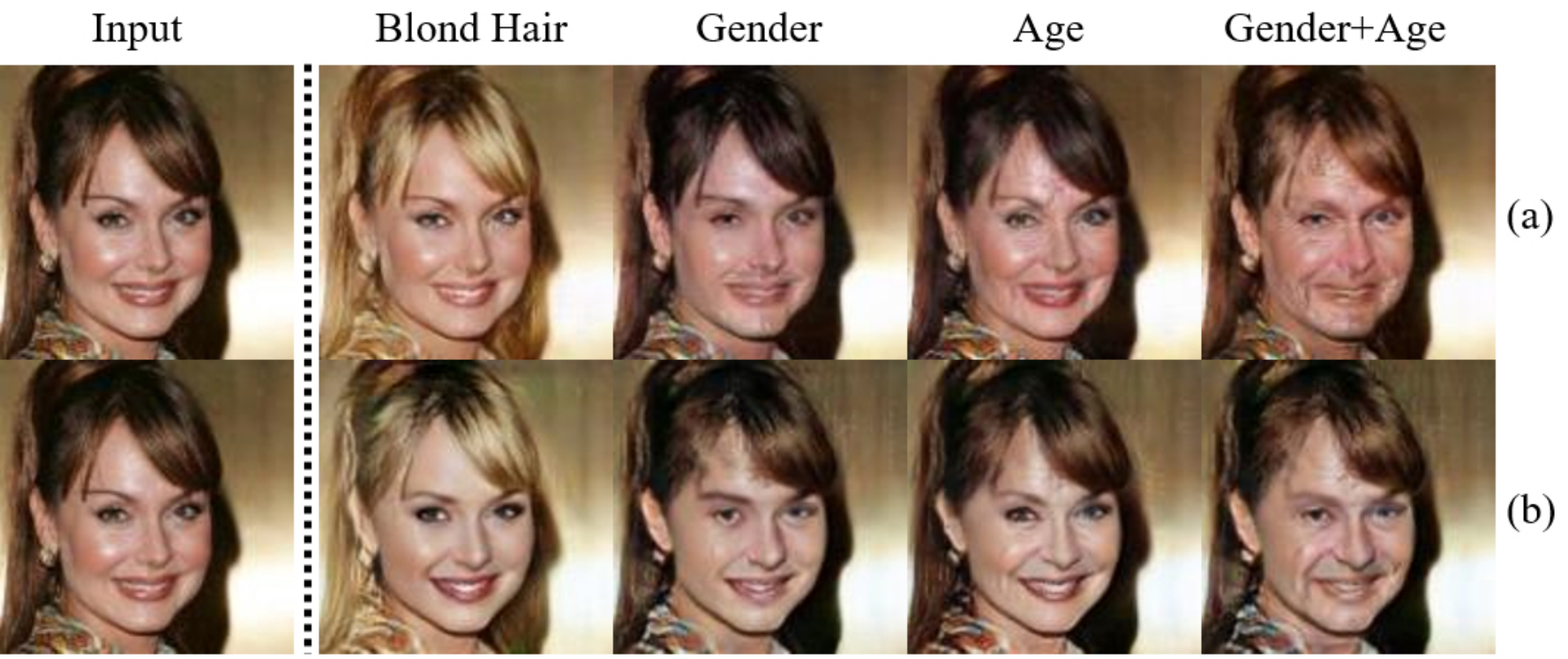}
  \caption{Analysis on tuning hyper-parameter for reconstruction loss ($\lambda_3$). (a) $\lambda_3$ = 20. (b) $\lambda_3$ = 1.}
  \label{fig_hyper1}\vspace{-0.3cm}
\end{figure}
\begin{figure}[t]
  \centering
  \includegraphics[width=0.48\textwidth]{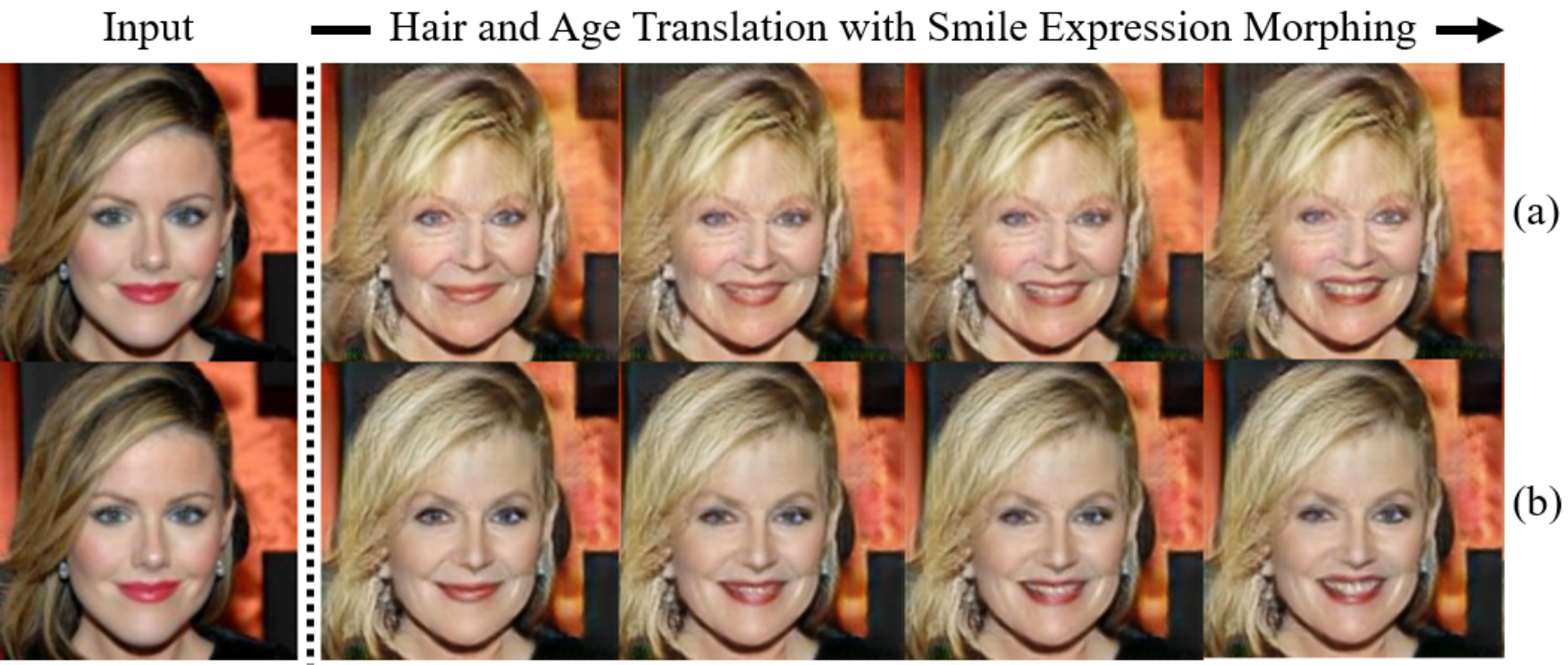}
  \caption{Analysis on tuning hyper-parameter for segmentation loss ($\lambda_2$). (a) $\lambda_2$ = 1. (b) $\lambda_2$ = 30.}
  \label{fig_hyper2}\vspace{-0.3cm}
\end{figure}

\noindent\textbf{Hyper-parameter analysis}
We provide additional results on hyper-parameter analysis to explain the trade-off in parameter setting to provide higher-quality results. As shown in Fig.~\ref{fig_hyper1} and Fig.~\ref{fig_hyper2}, increasing the weight of reconstruction loss tends to blur the output image, generate lower-quality results, but ensure the output images to be more similar to input images. On the other side, increasing the weight for segmentation loss tends to produce sharper, realistic output. However, increasing $\lambda_2$ too much will produce faces with more makeup (look younger) in \emph{Young2Old} translation. With lower $\lambda_2$, classification loss seems taking more effects that the generated results are older than~\ref{fig_hyper2}(b). In practical applications, these parameters can be tuned on demand.

\begin{figure}[t]
  \includegraphics[width=0.48\textwidth]{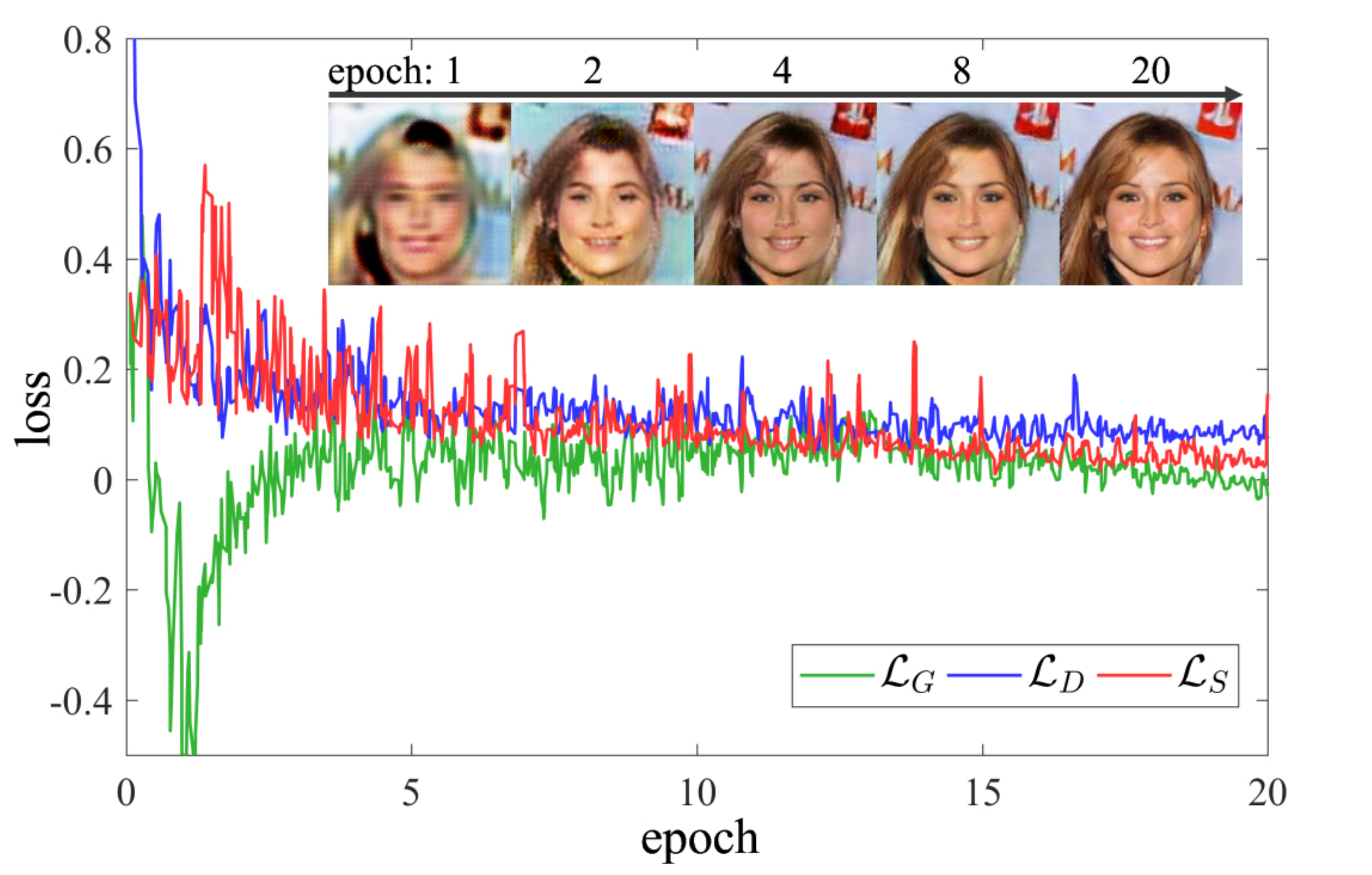}
  \caption{Model convergence. Losses of the generator, discriminator and segmentor are plotted together to show that our model converges fast and stably. With the networks converge, the output images of the generator are becoming sharper and more realistic with better consistency with the target attributes and segmentation.}
  \label{fig_losses}\vspace{-0.3cm}
\end{figure}

\noindent\textbf{Model convergence.~}
To demonstrate that SGGAN converge well with our introduced the segmentor and segmentation loss. The losses during the training process are plotted together with the corresponding generated results, as shown in Fig. \ref{fig_losses}. With of losses of $S$, $G$ and $D$ converging, the visual quality of generated results improves continuously.


\section{Conclusions}
In this paper, we have improved the multi-domain image translation problem by developing a Segmentation Guided Generative Adversarial Networks (SGGAN). Segmentation information is leveraged to provide strong regulations and guidance in image translation to avoid any ghost image or blurry detail. Moreover, this approach provides a spatial controllability called face morphing as an additional feature, which can align the input face images to the target segmentations and interpolate the intermediate faces from smiling to not smiling. We also discuss the proposed SGGAN model by providing an ablation study, a parameter analysis and a study of model convergence. Experimental results have demonstrated that the proposed SGGAN framework is effective and promising in face image translation applications.

\end{document}